\documentclass[twoside,leqno,twocolumn]{article}

\usepackage[letterpaper]{geometry}
\usepackage{graphicx}
\usepackage{ltexpprt}
\usepackage[colorlinks, linkcolor=black, anchorcolor=black, citecolor=black]{hyperref}
\usepackage{cite}
\usepackage{amsmath,amssymb,amsfonts}
\usepackage{algorithmic}
\usepackage{textcomp}
\usepackage{xcolor}
\usepackage{booktabs}
\usepackage{multirow}
\usepackage{subcaption}
\usepackage{flushend}
\usepackage{marvosym}
\usepackage{appendix}

\begin{document}

\title{\Large GE-AdvGAN: Improving the transferability of adversarial samples by gradient editing-based adversarial generative model}

\author{
  Zhiyu Zhu\thanks{Z. Zhu, H. Chen, Z. Jin and D. Yuan are with the School of Electrical and Computer Engineering, The University of Sydney, Australia (e-mail: \{zzhu2018, zjin0915\}@uni.sydney.edu.au), \{huaming.chen, dong.yuan\}@sydney.edu.au)} \and
  Huaming Chen\footnotemark[1] \textsuperscript{\Letter}\and
  Xinyi Wang\thanks{X. Wang is with the University of Malaya, Malaysia (e-mail: xinyiwangnoctis@outlook.com)}\and
  Jiayu Zhang\thanks{J. Zhang is with Suzhou Yierqi, China (e-mail: zjy@szyierqi.com)} \and
  Zhibo Jin\footnotemark[1] \and
  Kim-Kwang Raymond Choo\thanks{Prof. K.R. Choo is with the Department of Information Systems and Cyber Security, The University of Texas at San Antonio, USA (e-mail: Raymond.Choo@utsa.edu)} \and
  Jun Shen\thanks{Prof. J. Shen is with University of Wollongong, Australia (e-mail: jshen@uow.edu.au)} \and
  Dong Yuan\footnotemark[1] \textsuperscript{\Letter}
}

\date{}

\maketitle


\fancyfoot[R]{\scriptsize{Copyright \textcopyright\ 2024 by SIAM\\
Unauthorized reproduction of this article is prohibited}}





\begin{abstract} \small\baselineskip=9pt Adversarial generative models, such as Generative Adversarial Networks (GANs), are widely applied for generating various types of data, i.e., images, text, and audio. Accordingly, its promising performance has led to the GAN-based adversarial attack methods in the white-box and black-box attack scenarios. The importance of transferable black-box attacks lies in their ability to be effective across different models and settings, more closely aligning with real-world applications. However, it remains challenging to retain the performance in terms of transferable adversarial examples for such methods. Meanwhile, we observe that some enhanced gradient-based transferable adversarial attack algorithms require prolonged time for adversarial sample generation. Thus, in this work, we propose a novel algorithm named GE-AdvGAN to enhance the transferability of adversarial samples whilst improving the algorithm's efficiency. The main approach is via optimising the training process of the generator parameters. With the functional and characteristic similarity analysis, we introduce a novel gradient editing (GE) mechanism and verify its feasibility in generating transferable samples on various models. Moreover, by exploring the frequency domain information to determine the gradient editing direction, GE-AdvGAN can generate highly transferable adversarial samples while minimizing the execution time in comparison to the state-of-the-art transferable adversarial attack algorithms. The performance of GE-AdvGAN is comprehensively evaluated by large-scale experiments on different datasets, which results demonstrate the superiority of our algorithm. The code for our algorithm is available at: \hyperlink{https://github.com/LMBTough/GE-advGAN
}{https://github.com/LMBTough/GE-advGAN}.

\textbf{Keywords:} Gradient editing, Adversarial transferability, GAN-based adversarial attack, Computing optimization
\end{abstract}

\section{Introduction}

Adversarial Generative Models (AGM) have exhibited decent performance in generating various types of data such as images, text, and audio~\cite{song2017pixeldefend,ren2020generating,xie2021enabling}. They mainly consist of a generator and a discriminator, where the generator generates samples that resemble real data, while the discriminator attempts to differentiate the generated samples from real data. These two components are trained in an adversarial manner, ultimately ensuring that the generator can generate more realistic samples while the discriminator's judgments become more accurate. Based on the fundamental architecture, Generative Adversarial Networks~\cite{goodfellow2020generative} (GANs) represent a special manifestation of the process where the generator and discriminator compete with each other and are continuously optimised in the training process.

For adversarial attacks, they can be categorized into white-box and black-box attacks~\cite{sablayrolles2019white,papernot2017practical}. In white-box attacks, attackers have access to information about the victim model's structure and parameters, which assists in generating plausible adversarial samples. Differently, black-box attacks assume attackers having limited information about the model. 
It is noted that the transferable adversarial attacks involve the generation of high-quality adversarial samples by utilising a local surrogate model that closely resembles the victim model~\cite{long2022frequency,zhang2022improving,zhu2023improving}. It ensures the generated samples exhibit effective attack performance without querying the victim model~\cite{cheng2019improving}. 

To obtain high-quality adversarial samples~\cite{isola2017image},~\cite{xiao2018generating} proposed an adversarial attack algorithm for white-box and black-box attacks based on the vanilla GANs, called AdvGAN. Denoting the generator as $G$ to generate perturbations in AdvGAN in a white-box attacking environment, once $G$ is trained, there is no need to continuously access the victim model information. It resolves the requirement of multiple queries to the model to train optimal adversarial samples in conventional white-box attacks.
Furthermore, a dynamic distillation process is introduced in the discriminator (hereinafter denoted as $D$), allowing AdvGAN to be applicable to black-box attacks~\cite{hinton2015distilling}. The algorithm integrates the feed-forward and discriminator networks in a novel way to construct $G$ and $D$ for adversarial sample generation.

However, on the one hand, despite the promising results in black-box attacks, i.e., a 92.76\% attack success rate on the MNIST Adversarial Examples Challenge~\cite{madry2017towards}, the attack success rate will be impacted by the adversarial defenses against queries. 
On the other hand, GAN-based adversarial attack methods have neglected the potential of transferable adversarial attacks, which require less data preparation but can be broadly applied for query-based attacks~\cite{qin2021random}.
Considering the advantages of transferable adversarial attacks, we discuss the feasibility of GAN-based adversarial attack methods to generate highly transferable adversarial samples, which are functionally and characteristically similar (see discussion in Section.~\ref{relatedwork}) for the source model and victim model in transferable adversarial attacks.

Though transferable adversarial attacks can assist in the attack success rate when facing adversarially trained models and black-box environments, it is limited to extending their performance to a wider landscape. The attack success rate of transferable samples is highly impacted by the differences between the target model and the local source model. Another overfitting issue~\cite{xie2022improving} that arises in local surrogate models during white-box training also has impacts on the success rate. Moreover, the existing gradient-based transferable adversarial attack methods, such as NIM~\cite{lin2019nesterov}, MIM~\cite{dong2018boosting}, and VMI-FGSM~\cite{wang2021enhancing}, require gradient information calculation for the input samples, which dramatically increase computation time in large-scale datasets. 

Therefore, we are motivated to \textit{enhance the transferability of the adversarial samples} and \textit{the computational efficiency}. By exploring the necessity and feasibility of GAN-based adversarial attack methods for generating transferable samples, we find that the gradient update process of $G$ can be edited in conjunction with a transferable approach, which is similar to training high transferable samples on a local surrogate model. Recent studies have shown that DNNs have different sensitivities to different frequency domains in the presence of human-added perturbations~\cite{yin2019fourier}. Performing spectral transformations on the inputs to explore frequency information can provide new insights for model enhancement and are critical in generating transferable samples~\cite{long2022frequency}. To fulfill the aforementioned motivations, we perform frequency-based exploration as the basis for gradient editing direction to generate highly transferable adversarial samples. Moreover, since no additional gradient computation is required once $G$ has been trained, our method has a faster attack speed compared to other gradient-based transferable attack algorithms.

In this work, we further investigate the GAN-based adversarial attack with transferability, in which we propose a novel method named GE-AdvGAN to utilize the gradient editing mechanism from the state-of-the-art transferable attack algorithms and optimize the parameter updating process of $G$ (called $GE-G$) in the GAN models. Fig.~\ref{fig:flow} illustrates the schematic diagram of our algorithm. To explore the advantages of GE-AdvGAN in terms of transferability and efficiency compared to other gradient-based transferable adversarial attacks, we conduct the experiments in various white-box and black-box settings with the performance metrics of the attack success rates and execution times.
We summarise the key contributions of this paper as follows:



\begin{figure}
    \centering
    \includegraphics[width=\linewidth]{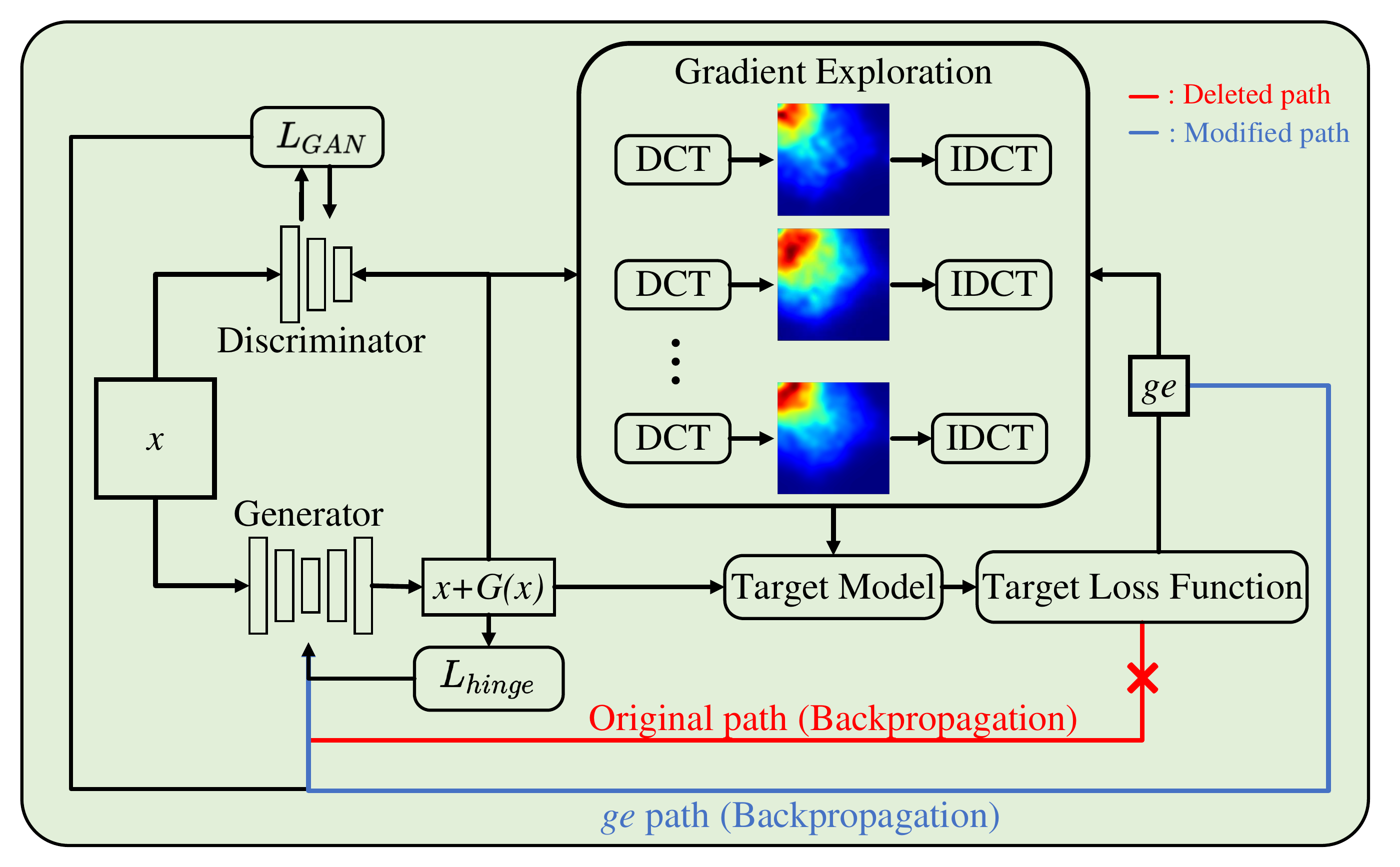}
    \caption{GE-AdvGAN Schematic Diagram (the red line represents the original path, we use the blue $ge$ path instead of the original path.)}
    \label{fig:flow}
\end{figure}


\begin{itemize}
    \item We propose a novel gradient editing algorithm, named GE-AdvGAN, to optimize the gradient training process of AGM.
    \item Based on the model sensitivity to different frequency ranges, we, for the first time, have incorporated the frequency-based exploration as the basis for gradient editing directions in GE-AdvGAN.
    \item We compare the performance of GE-AdvGAN with other gradient-based transferable adversarial attack algorithms regarding the transferability of adversarial samples and algorithmic run-time cost. 
    \item We release the replication package of GE-AdvGAN for further research and development.
\end{itemize}

\section{Related work}\label{relatedwork}
\subsection{GAN-based adversarial attacks}
Normally, GAN-based adversarial algorithms include a generator network and a discriminator network. The generator network generates synthetic samples, while the discriminator network aims to distinguish between real and synthetic samples. Some literature algorithm examples are DCGAN~\cite{radford2015unsupervised}, WGAN~\cite{arjovsky2017wasserstein}, cGAN~\cite{mirza2014conditional} and CycleGAN~\cite{zhu2017unpaired}. These have been extensively developed in the field of images, such as in style transfer applications. 

By delving into the details of these GAN-based algorithms, we find that they exhibit high flexibility and optimisation potential for both $G$ and $D$ components. It motivates us to implement the gradient editing optimisation methods on the generator of $G$. Recalling the discussion of AdvGAN for the functional and characteristic similarity, hereafter, we discuss the details in the context of the source model and victim model in transferable adversarial attacks.

\paragraph{Functional Similarity}
Essentially, the objective of $G$ is to \textit{generate} adversarial samples with added perturbations, then feed them into $D$ for \textit{distinction detection}, ensuring that the adversarial samples resemble benign samples as closely as possible, ultimately leading to misleading decisions by the deep neural network. This ingenious attack process provides a hypothesis that considering transferable adversarial attacks, the \textit{generation} of adversarial samples on the local surrogate model and the \textit{transferability detection} on the victim model are functionally similar to the $G$ and $D$ architectures in AdvGAN. If $G$ can both fulfill the function of generating adversarial samples and possess transferability detection capability (aiming to ensure the generated samples are transferable attack samples), then the samples evaluated by $D$ (aiming to preserve minimal changes from the original samples) are highly likely to be applicable in the context of transferable adversarial attacks.
\paragraph{Characteristic Similarity}
From characteristic perspective, both $G$ and $D$, as well as the source and victim models, exhibit independence and flexibility for editing. In AdvGAN, the feed-forward network and discriminator network are mutually independent and can be adjusted through editing. Similarly, in transferable adversarial attacks, although it requires a certain degree of consistency between the source and victim models, the generation of adversarial samples and the evaluation of transferability are independent. Moreover, attackers have the flexibility to select different source models for training depending on the chosen victim models. It offers a high degree of adaptability in model selection. Thus, with the functional and characteristic similarities, we aim to optimise the vanilla AdvGAN by incorporating the principles of transferable adversarial attacks.
\subsection{Transferable adversarial attacks}

By leveraging different gradient techniques, some algorithms can generate highly transferable adversarial samples. For example, MIM~\cite{dong2018boosting} is an iterative attack method that introduces a momentum term in each iteration to extend the search space and further use model gradients to enhance the perturbation of adversarial samples. NIM~\cite{lin2019nesterov} incorporates the concept of momentum and introduces a prediction step in the gradient update process to improve the convergence speed of the optimization algorithm. VMI-FGSM~\cite{wang2021enhancing}, on the other hand, constructs a variance tuning technique to preserve the gradient variance of the previous iteration to adjust the current gradient. Alternative transferable adversarial attack algorithms utilising input transformation~\cite{dong2019evading,gao2020patch} and feature-level attacks~\cite{wang2021feature,zhang2022improving,jin2023danaa} can also generate adversarial samples via sample transformation and estimating the importance of intermediate layer neurons.

In the literature, we note that transferable black-box algorithms play a significant role in adversarial attacks. However, generating effective attack samples often requires lots of gradient-related computations, which computation efficiency needs to be improved. 
\subsection{Frequency-based analysis in adversarial attacks}
Several studies have highlighted the significant relationship between DNNs and the frequency domain~\cite{wang2020high,guo2018low,yin2019fourier}. DNNs can effectively capture high-frequency information within images that is imperceptible to human eye~\cite{wang2020high}. In the context of adversarial attacks, DNNs display varying degrees of sensitivity to perturbations introduced across different frequency domains, with high-frequency regions demonstrating heightened susceptibility~\cite{yin2019fourier}.

To investigate the perturbation impacts on the low-frequency regions of images on target models,~\cite{guo2018low} introduces an adversarial algorithm that exclusively targets the low-frequency domain. Their findings underscore the significance of the low-frequency domain in influencing DNN predictions. In order to generate highly transferable adversarial attack samples,~\cite{long2022frequency} incorporates spectral transformation to the input to perform model augmentation in the frequency domain. With the insights from existing literature, we argue that the frequency domain holds valuable information for adversarial attacks, and its strategic utilization can effectively enhance the transferability of adversarial samples.


\section{Preliminaries}\label{preliminaries}
To comprehensively discuss our approach in Sec.~\ref{approach}, we firstly define the adversarial attack and introduce a spectral transformation method of Discrete Cosine Transformation (DCT) for our gradient editing process in the GE-AdvGAN algorithm.

\subsection{Problem definition}
Formally, let us consider a clean feature space where benign samples are represented by $X \in  \mathbb{R}^{W\times H\times C}$. Here, $W$ denotes the image width, $H$ represents the image height, and $C$ is the number of channels. Let $Y$ be the set consisting of $S$ different labels. For a sample $(x_i, y_i)$ in the data space, where $x_i \subseteq X$ and $y_i \subseteq Y$, the objective of a deep neural network is to learn a classifier $f: X \rightarrow Y$. In a normal classification task, the network is typically able to correctly classify the samples. Specifically, for an input sample $x_i$ and its corresponding label $y_i$, the classification result can be defined as $\hat{y_i} = f(x_i)$.

However, for adversarial attack, the goal of the adversary is to deceive the network $f$ by introducing imperceptible perturbations to manipulate the input samples $\delta \in \mathbb{R}^{W\times H\times C}$. Let us consider the feature space containing adversarial samples as $x_{adv}=x + \delta$. Then the classification result in an untargeted adversary attack satisfies $f(x_{adv})\ne y_{i}$, where $y_{i}$ is the true label.

Note that in GAN-based adversarial algorithms, the perturbation $\delta$ is typically generated by the generator $G$. Therefore, in order to generate an adversarial example capable of fooling the target network $f$, $G$ will be incorporated in the following mathematical formula:
\begin{equation}
\small
    x_{adv}=x_{i}+G(x_{i})
\end{equation}
such that in an untargeted attack
\begin{equation}
\small
    f(x_{adv})\neq y_{i}
\end{equation}
\begin{equation}
\small
    \left \| x_{adv}-x \right \|_{p}\le  \epsilon 
\end{equation}
where $\left \| \cdot \right \|_{p} $ represents the $n$-order norm(e.g., $Lp$ norm), $\epsilon$ denotes the maximum perturbation.

\subsection{Discrete Cosine Transformation}
In~\cite{ahmed1974discrete}, a novel digital processing technique called DCT is proposed to improve the comprehension of pattern recognition tasks. It is important to emphasize that DCT can be employed in the image transformation process, particularly for converting an image from the spatial domain to the frequency domain. This conversion facilitated by DCT allows for evaluating the sensitivity of different regions to adversarial attacks. The process of DCT can be expressed as:
\begin{equation}
\small
    \begin{aligned}
\mathcal{D}(x)_{[u, v]}= & \frac{1}{\sqrt{2 N}} C(u) C(v) \sum_{x=0}^{N-1} \sum_{y=0}^{N-1} x[k, m] \\
& \cos \left[\frac{(2 k+1) i \pi}{2 N}\right] \cos \left[\frac{(2 m+1) j \pi}{2 N}\right]
    \end{aligned}
\end{equation}
The inverse discrete cosine transformation (IDCT) functions as the inverse of DCT, allowing the transformation of the image back to the spatial domain. It is important to note that both DCT and IDCT operations are lossless, and can facilitate the gradient calculations~\cite{ahmed1974discrete}.

DCT provides an alternative view on adversarial attacks via leveraging the frequency domain. Several studies showcase improved results in transferable black-box attacks by exploiting frequency domain techniques~\cite{duan2021advdrop,long2022frequency}. We find that the frequency domain can enhance the consistency of spatial domain attacks, thereby effectively guiding the attack direction. This characteristic offers insights to determine the attack direction for gradient editing, which will be discussed in Sec.~\ref{gt}.
\section{Method}\label{approach}
In this section, we introduce and demonstrate the mathematical principles and feasibility of applying gradient editing in $G$. Then, we elaborate on the specific implementation approach of GE-AdvGAN.

\subsection{Gradient editing based on frequency domain exploration}\label{gt}
\subsubsection{Object selection of gradient editing} According to the discussion of the AdvGAN loss function in~\cite{xiao2018generating} (see \textbf{appendix} for details in our GitHub), we decompose the loss function into three distinct components $L_{adv}^{f}$, $ L_{GAN}$ and $ L_{hinge}$, using $\alpha$ and $\beta$ to adjust the importance of these three parts. To independently train $G$ and $D$, we need to fix the parameters of $D$ while training $G$, and vice versa. Based on this consideration, we derive the following parameter update equations for $G$ and $D$:
\begin{equation}
\small
\label{eqWGoriginal}
    W_{G}=W_{G}-\eta \left ( \frac{\partial L_{adv}^{f}}{\partial W_{G}}+\alpha \frac{\partial L_{GAN}}{\partial W_{G}}+\beta \frac{\partial L_{hinge}}{\partial W_{G}}  \right ) 
\end{equation}
\begin{equation}
\small
    W_{D}=W_{D}-\eta \left ( \frac{\partial L_{GAN}}{\partial W_{D}}  \right ) 
\end{equation}
It is worth noting that since the primary objective of $D$ is to control the authenticity of the perturbations generated by $G$, it does not assess how the perturbations affect the final adversarial outcome. Therefore, our gradient editing algorithm does not require any modifications to $D$.

\subsubsection{Gradient extension} Now let us consider the parameter update process for $G$ in Eq.~\ref{eqWGoriginal}. It can be observed that the gradient of $G$ consists of $\frac{\partial L_{adv}^{f}}{\partial W_{G}}$, $\alpha \frac{\partial L_{GAN}}{\partial W_{G}}$, and $\beta \frac{\partial L_{hinge}}{\partial W_{G}}$. Among them, $\alpha \frac{\partial L_{GAN}}{\partial W_{G}}$ corresponds to the authenticity constraint imposed by $D$, and $\beta \frac{\partial L_{hinge}}{\partial W_{G}}$ corresponds to the magnitude constraint on the generated perturbations. It is evident that $\frac{\partial L_{adv}^{f}}{\partial W_{G}}$ is directly related to the adversarial effects.

By applying the chain rule, we can expand the gradient propagation process of Eq.~\ref{eqWGoriginal} as follows:
\begin{equation}
\small
\label{eq333}
\nabla_{W_{G}} L_{adv}^{f}=\frac{\partial L_{adv}^{f}}{\partial (x+G(x))} \cdot \frac{\partial (x+G(x))}{\partial G(x)} \cdot \frac{\partial G(x)}{\partial W_{G}}
\end{equation}
such that:
\begin{equation}
\small
\label{eqtriangle}
        W_{G}=W_{G}-\eta \left ( \nabla_{W_{G}} L_{adv}^{f}+\alpha \frac{\partial L_{GAN}}{\partial W_{G}}+\beta \frac{\partial L_{hinge}}{\partial W_{G}}  \right ) 
\end{equation}
In this context, $\frac{\partial (x+G(x))}{\partial G(x)}=1$, so it can be omitted. $\frac{\partial L_{adv}^{f}}{\partial (x+G(x))}$ represents the change in the adversarial sample $x+G(x)$ with respect to the variation in the loss function. In other words, the degree of change in the adversarial sample and its corresponding aggressiveness are determined by $\frac{\partial L_{adv}^{f}}{\partial (x+G(x))}$. On the other hand, $\frac{\partial G(x)}{\partial W_{G}}$ corresponds to the update process of the parameters in $G$ and their impact on the adversarial sample.

\subsubsection{Feasibility analysis of gradient editing} 
Based on the fundamental principles of calculus, we can establish the following equivalence relations.
\begin{equation}
\small
\label{equal}
    \frac{\partial L_{adv}^{f}}{\partial (x+G(x))}=\frac{\partial L_{adv}^{f}}{\partial x}=\frac{\partial L_{adv}^{f}}{\partial G(x)}
\end{equation}
where Eq.~\ref{equal} corresponds to the process of attack as we discussed in Eq.~\ref{eq333}, which can be understood as having three aspects of influence: (i) Overall variation, (ii) Variation in $x$, and (iii) Variation in $G(x)$. 

The overall variation in $x+G(x)$ represents the update process of the original GAN-based adversarial algorithm. If we interpret this process as (ii), then as shown in Eq.~\ref{fgsm}, the variation in $x$ leads to the generation of attack, i.e., the origin of attack in the FGSM~\cite{goodfellow2014explaining} training process.
\begin{equation}
\small
\label{fgsm}
    x=x+\eta \cdot sign(\frac{\partial L_{adv}^{f}}{\partial x})
\end{equation}
Furthermore, for (iii), it can be understood as equivalent to the case of (i). It is worth noting that the variations in (i) and (iii) are primarily caused by changes in the parameters of $G$, while the variation in (ii) is primarily derived from the changes in the sample $x$ itself.

To sum up, we can enhance the transferability by modifying the gradient propagation of $x+G(x)$. Inspired by the work of Long et al.~\cite{long2022frequency}, we realize that different models exhibit different sensitivities to various frequency domains, even when they are trained on similar or identical datasets. So, if (ii) can explore the gradient ascent based on different frequency ranges of $x$, we can incorporate this effect into the training process of $G$ through the equivalent conditions in Eq.~\ref{equal}.

\subsubsection{Frequency domain exploration} To explore different frequency domains of the input samples $x$, we first use Discrete Cosine Transform (DCT) to map the samples into the frequency space. Then, we generate $N$ approximate samples $x_{f_i}$ of $x+G(x)$ by adding noise to the original samples and applying random transformations in the frequency space. $\sigma$ is the variance used in frequency domain exploration. The process is mathematically represented as:
\begin{equation}
\small
    x_{f_i}=IDCT(DCT(x+G(x)+N(0,I)\cdot \frac{\epsilon}{255})\ast N(1,\sigma))
\end{equation}
We want $G$ to be sensitive to the approximate samples $x_{f_i}$ throughout the training process. Based on these considerations, we propose a novel gradient editing approach as follows:
\begin{equation}
\small
\label{ge}
    ge=-sign\left ( \frac{1}{N}\sum_{i=1}^{N}\frac{\partial L(x_{f_i},y)}{\partial x_{f_i}}   \right ) 
\end{equation}
where $ge$ represents the target gradient that needs to be modified. We randomly select $N$ approximate samples $x_{f_i}$ and average the results. It is important to note that $x_{f_i}$ must be generated iteratively in real-time during the training process and cannot be pre-generated instead. Additionally, since the original samples $x$ require gradient ascent to achieve the adversarial objective, while the $G$ training process employs gradient descent, we need to add a negative sign at the beginning to ensure consistency.
\subsection{GE-AdvGAN}

After incorporating Eq.~\ref{eqtriangle},~\ref{equal} and~\ref{ge}, we obtain the parameter update process for $GE-G$:
\begin{equation}
\small
\nabla_{W_{G}} L_{adv}^{f}=ge \cdot \frac{\partial (x+G(x))}{\partial G(x)} \cdot \frac{\partial G(x)}{\partial W_{G}}
\end{equation}
such that:

\begin{equation}
\small
\label{WGEG}
W_{G}=W_{G}-\eta \left ( \nabla_{W_{G}} L_{adv}^{f}+\alpha \frac{\partial L_{GAN}}{\partial W_{G}}+\beta \frac{\partial L_{hinge}}{\partial W_{G}}  \right ) 
\end{equation}

For the training process, there are two steps for gradient editing. The first step involves editing the gradient graph during real-time gradient propagation. The second step involves truncating $\frac{\partial L_{adv}^{f}}{\partial (x+G(x))}$ and replacing it with another target. This is because, as mentioned earlier in Eq.~\ref{equal}, we know that the three parts are equivalent, and thus, it is necessary to apply the (ii) part, Variation in $x$, to the (i) part, overall variation.

Since editing the gradient graph requires GPU optimization, we adopt a truncation method for the sake of generality. Therefore, to obtain the final GE-AdvGAN, we replace $L_{adv}^{f}$ with the following expression:
\begin{equation}
\small
\label{eqB}
    L_{adv}^{f}=-\frac{1}{B} sum([x+G(x)]\ast ge)
\end{equation}
where $B$ denotes the batch size. The gradient computation for Eq.~\ref{eqB} is equivalent to the derivation result of Eq.~\ref{WGEG}, and it can be easily applied to the actual calculation process. During the training process, we only need to replace the $L_{adv}^f$ in AdvGAN to complete the training. The complete flowchart of our algorithm is presented in Fig.~\ref{fig:flow}.
\section{Experiments}
In this section, we empirically evaluate the performance of GE-AdvGAN over other state-of-the-art methods in terms of the transferability of generated adversarial samples. Additionally, we investigate the computation efficiency improvement of GE-AdvGAN compared to other baselines.
\subsection{Experimental setup}
\subsubsection{Dataset} In order to ensure the fairness of experiment evaluation, we employ the same dataset selection method as in~\cite{zhang2022improving}. The dataset consists of 1000 randomly selected images from the ILSVRC 2012 validation set~\cite{russakovsky2015imagenet}, encompassing various categories.

\subsubsection{Models} We utilise four widely adopted models in image classification tasks, namely Inception-v3~\cite{szegedy2016rethinking}, Inception-v4~\cite{szegedy2017inception}, Inception-ResNet-v2~\cite{szegedy2017inception}, and ResNet-v2-152~\cite{he2016deep}, as the source models to compare the attack performance of our algorithm and competing methods on seven different models. Among them, Inception-v3, Inception-v4, Inception-ResNet-v2, and ResNet-152 are models without any defensive training. On the other hand, the ensemble of three adversarially trained Inception-v3 (Inception-v3-ens3), the ensemble of four adversarially trained Inception-v3 (Inception-v3-ens4), and the ensemble of three adversarially trained Inception-ResNet-v2 (IncRes-v2-ens3) are more complicated models that incorporate the adversarial defenses techniques.

\subsubsection{Metrics} 
In the experiment, we apply the attack success rate (ASR) to evaluate our algorithm. ASR measures the proportion of samples that successfully mislead the model and cause misclassification among the entire dataset. A higher ASR indicates better attack performance of a method. 
ASR is computed as follows:
Stochastic Variance Reduced Ensemble Adversarial Attack for Boosting the Adversarial Transferability
\begin{equation}
\small
    ASR=\frac{Number\ of\ misleading\ samples}{Number\ of\ total\ samples} 
\end{equation}
To evaluate the authenticity of adversarial examples, we employ the perturbation ratio (PR) as described in~\cite{zhang2021survey}. A lower PR indicates that the attacked images are closer to the original ones. We measure PR in two ways: using the absolute value, which we refer to as the Absolute Perturbation Ratio (APR), and through square calculation, termed as the Square Perturbation Ratio (SPR).
APR is computed as follows:
\begin{equation}
\small
    APR=\frac{1}{N} \ast \frac{abs(x'-x)}{W\cdot H\cdot C} 
\end{equation}

SPR is computed as follows:
\begin{equation}
\small
    SPR=\frac{1}{N} \ast \frac{(x'-x)^2}{W\cdot H\cdot C} 
\end{equation}

where $N$ represents the total number of samples in the dataset, $x'$ denotes the attacked example, $x$ represents the original image, $W$ denotes the image width, $H$ represents the image height, and the number of channels is C. Additionally, we use Frames Per Second (FPS) as an evaluation metric for our running efficiency. It is important to note, since AdvGAN requires only a single training session to conduct attacks and does not need further training afterward, our time measurement focuses solely on the interaction time.
\begin{equation}
\small
    FPS=\frac{Number\ of\ samples}{Running\ time\ of\ these\ samples} 
\end{equation}

\subsubsection{Baseline attack algorithms} We select six widely-used attack methods as competing algorithms, namely AdvGAN~\cite{xiao2018generating}, MIM~\cite{dong2018boosting}, NRDM~\cite{naseer2018task}, FDA~\cite{ganeshan2019fda}, FIA~\cite{wang2021feature}, and NAA~\cite{zhang2022improving}.

\subsubsection{Parameters} The source models used in this experiment include Inception-v3, Inception-v4, Inception-ResNet-v2, and ResNet-152, in toal four models. The parameter setting is consistent for these four models: \textit{adv lambda} is set to 10, $N$ is 10, \textit{epsilon} ($\epsilon$) is set to 16, Epoch is 60, Change threshold is between [20, 40], Discriminator ranges is between [1, 1], and Discriminator learning rate is set to [0.0001, 0.0001]. The parameters that differ among these models are as follows: when Inception-v3 is the source model, \textit{sigma} ($\sigma$) is set to 0.7, while for the other three models, \textit{sigma} is set to 0.5. When Inception-ResNet-v2 is the source model, the Generator ranges and Generator learning rate are set to [1, 1] and [0.0001, 0.000001], respectively, whereas for the other three models, they are set to [2, 1] and [0.0001, 0.0001], respectively.
\subsection{Experimental results}

\begin{table}[htpb]
\centering
\caption{FPS of multiple methods}
\label{tab:FPS}
\resizebox{0.35\textwidth}{!}{%
\begin{tabular}{@{}c|cccccc@{}}
\toprule
    & NAA & FIA & MIM & NRDM & FDA  & GE-AdvGAN \\ \midrule
FPS & 1.3 & 1.8 & 7.4 & 10.8 & 13.4 & \textbf{2217.7}    \\ \bottomrule
\end{tabular}%
}
\end{table}

\begin{table}[htpb]
\centering
\caption{Attack success rate of multiple methods on different models}
\label{tab:asr-results}
\resizebox{0.5\textwidth}{!}{%
\begin{tabular}{@{}c|c|ccccccc@{}}
\toprule
Source Model               & Method    & Inc-v3        & Inc-v4        & IncRes-v2     & Res-152       & Inc-v3-Ens3   & Inc-v3-Ens4   & IncRes-v2-Ens \\ \midrule
\multirow{7}{*}{Inc-v3}    & AdvGAN    & 54.7          & 36.5          & 15.8          & 48.6          & 48.8          & 48.8          & 26.9          \\
                           & MIM       & \textbf{100}           & 41.9          & 39.7          & 32.8          & 14.9          & 15.7          & 8.2           \\
                           & NRDM      & 90.4          & 61.4          & 52.5          & 49.9          & 9.5           & 12.9          & 4.7           \\
                           & FDA       & 82            & 42.9          & 37.1          & 35.1          & 9.3           & 12.2          & 5             \\
                           & FIA       & 97            & 79.1          & 77.8          & 71.8          & 43.1          & 44.2          & 23.2          \\
                           & NAA       & 97            & 83            & \textbf{80.6} & 74.7          & 49.5          & 50.4          & 31.5          \\
                           & GE-AdvGAN & 95.9          & \textbf{90.9} & 75.1          & \textbf{88.1} & \textbf{82.4}          & \textbf{79.7}          & \textbf{69.9}          \\ \midrule
\multirow{7}{*}{Inc-v4}    & AdvGAN    & 65.2          & 74.3          & 9.1           & 64.9          & 20.9          & 54.3          & 24.1          \\
                           & MIM       & 58.2          & \textbf{99.9} & 45            & 40.4          & 17.7          & 20.3          & 9.7           \\
                           & NRDM      & 78            & 96.4          & 62.8          & 62.3          & 17.3          & 16.6          & 6.8           \\
                           & FDA       & 84.6          & 99.6          & 71.8          & 68.8          & 17.4          & 17.1          & 7             \\
                           & FIA       & 74.6          & 91            & 69.6          & 65.7          & 39.3          & 39.9          & 23.5          \\
                           & NAA       & 83.3          & 95.8          & \textbf{77.9} & 73.3          & 48            & 46.5          & 31.4          \\
                           & GE-AdvGAN & \textbf{88.4} & 95            & 69.1          & \textbf{81.4} & \textbf{81}   & \textbf{74.1} & \textbf{68.6} \\ \midrule
\multirow{7}{*}{IncRes-v2} & AdvGAN    & 37.8          & 33.9          & 33.4          & 28.2          & 11.7          & 14.1          & 12.1          \\
                           & MIM       & 60            & 51.9          & \textbf{99.2} & 42.2          & 21.7          & 23.3          & 12.3          \\
                           & NRDM      & 72.8          & 67.9          & 77.9          & 59.7          & 16.4          & 17.1          & 7.3           \\
                           & FDA       & 69            & 68            & 78.2          & 56.2          & 16.2          & 15.4          & 7.7           \\
                           & FIA       & 71            & 68.2          & 78.8          & 63.9          & 47.4          & 45.8          & 37.6          \\
                           & NAA       & 79.5          & 76.4          & 89.3          & 71.1          & 56.9          & 55            & 47.3          \\
                           & GE-AdvGAN & \textbf{87.4} & \textbf{83.4} & 90.6          & \textbf{80.3} & \textbf{72.1} & \textbf{63.3} & \textbf{57.1} \\ \midrule
\multirow{7}{*}{Res-152}   & AdvGAN    & 44.2          & 38            & 21.1          & 34.3          & 16.1          & 24.9          & 12.4          \\
                           & MIM       & 52.9          & 47.3          & 44.9          & 25.1          & 24.3          & 24.4          & 13.3          \\
                           & NRDM      & 72.7          & 68.8          & 59.5          & 89.9          & 20.3          & 18.1          & 9.3           \\
                           & FDA       & 15.7          & 9.2           & 8.3           & 26.2          & 9.3           & 9.7           & 4             \\
                           & FIA       & 80.7          & 78.2          & 77.5          & \textbf{98}   & 53            & 48.4          & 34.4          \\
                           & NAA       & 84.7          & \textbf{83.5} & \textbf{82.3} & 97.6          & 59.1          & 58.1          & 46.1          \\
                           & GE-AdvGAN & \textbf{85}   & \textbf{83.5} & 68.7          & 83.5          & \textbf{77.9} & \textbf{76.2} & \textbf{67.4} \\ \bottomrule
\end{tabular}%
}
\end{table}

All experiments in this study are conducted using the Nvidia RTX 2080ti. The specific experimental results are shown in Tab.~\ref{tab:asr-results}, which demonstrate the best performance of our method in comparison to other approaches across almost all models. Particularly, our method exhibits significant performance advantages, especially when applied to models that have implemented adversarial training methods, achieving state-of-the-art results. Specifically, when compared to the baseline model AdvGAN, our method demonstrates comprehensive superiority by achieving better adversarial attack transferability with lower perturbation rates. In comparison to several other adversarial attack methods, our approach shows substantial improvements in adversarial attack transferability performance. In comparison to competing algorithms, our method achieves an average Automatic Speech Recognition (ASR) improvement of 34\%, surpassing AdvGAN with an average ASR improvement of 46.3\%, and outperforming the state-of-the-art method of NAA~\cite{zhang2022improving} in the comparative set with an average improvement of 12.2\%. Notably, our method exhibits enhanced transferability on three complicated models that implement adversarial training, with an average increase in attack success rate of 26.9\% compared to NAA.

Furthermore, as shown in Tab.~\ref{tab:pert}, our method achieves better attack effectiveness with lower perturbation rates compared to AdvGAN. Specifically, our method achieves a higher average ASR improvement of 1.7-3.1 times at lower perturbation rates compared to AdvGAN. This indicates that our attack examples have lower levels of perturbation rate for the original images, while still achieving superior attacking performance.

\begin{table}[htpb]
\centering
\caption{Perturbation ratio and corresponding average attack success rate results}
\label{tab:pert}
\resizebox{0.45\textwidth}{!}{%
\begin{tabular}{@{}c|c|ccc@{}}
\toprule
Source Model               & Method    & SPR   & APR  & Average ASR \\ \midrule
\multirow{2}{*}{Inc-v3}    & AdvGAN    & 240.0 & 15.2 & 40.0        \\
                           & GE-AdvGAN & 212.1 & 13.9 & \textbf{81.2}        \\ \midrule
\multirow{2}{*}{Inc-v4}    & AdvGAN    & 241.7 & 15.3 & 44.7        \\
                           & GE-AdvGAN & 210.2 & 13.8 & \textbf{79.7}        \\ \midrule
\multirow{2}{*}{IncRes-v2} & AdvGAN    & 242.5 & 15.3 & 24.5        \\
                           & GE-AdvGAN & 210.8 & 13.8 & \textbf{76.3}        \\ \midrule
\multirow{2}{*}{Res-152}   & AdvGAN    & 235.0 & 15.0 & 27.3        \\
                           & GE-AdvGAN & 213.6 & 14.0 & \textbf{77.5}        \\ \bottomrule
\end{tabular}%
}
\end{table}

As shown in Tab.~\ref{tab:FPS}, we also evaluate the computational efficiency of our method in comparison to different approaches. Through these tests, our method demonstrates a significant advantage, leading in Frames Per Second (FPS) compared to all competing methods.

\subsection{Ablation study}
\begin{figure*}
\centering

\begin{subfigure}{0.23\textwidth}
  \includegraphics[width=\linewidth]{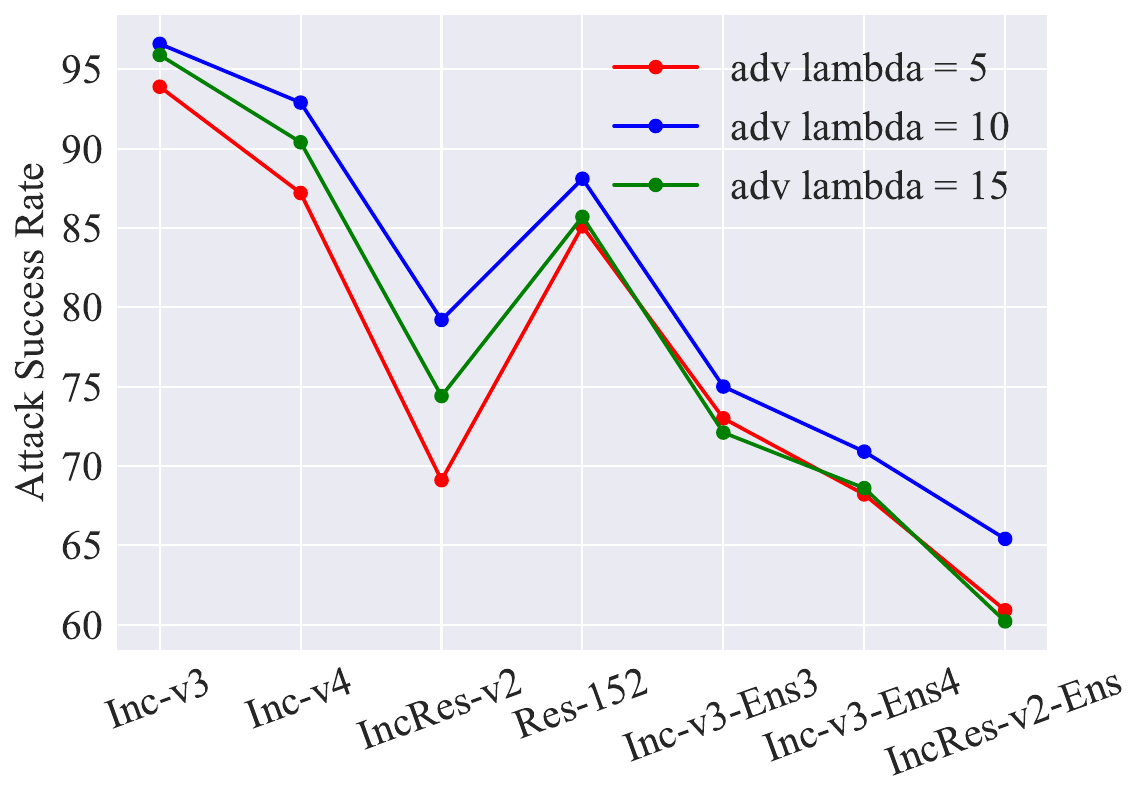}
  \caption{different \textit{adv lambda}}
  \label{fig:adv}
\end{subfigure}
\begin{subfigure}{0.23\textwidth}
  \includegraphics[width=\linewidth]{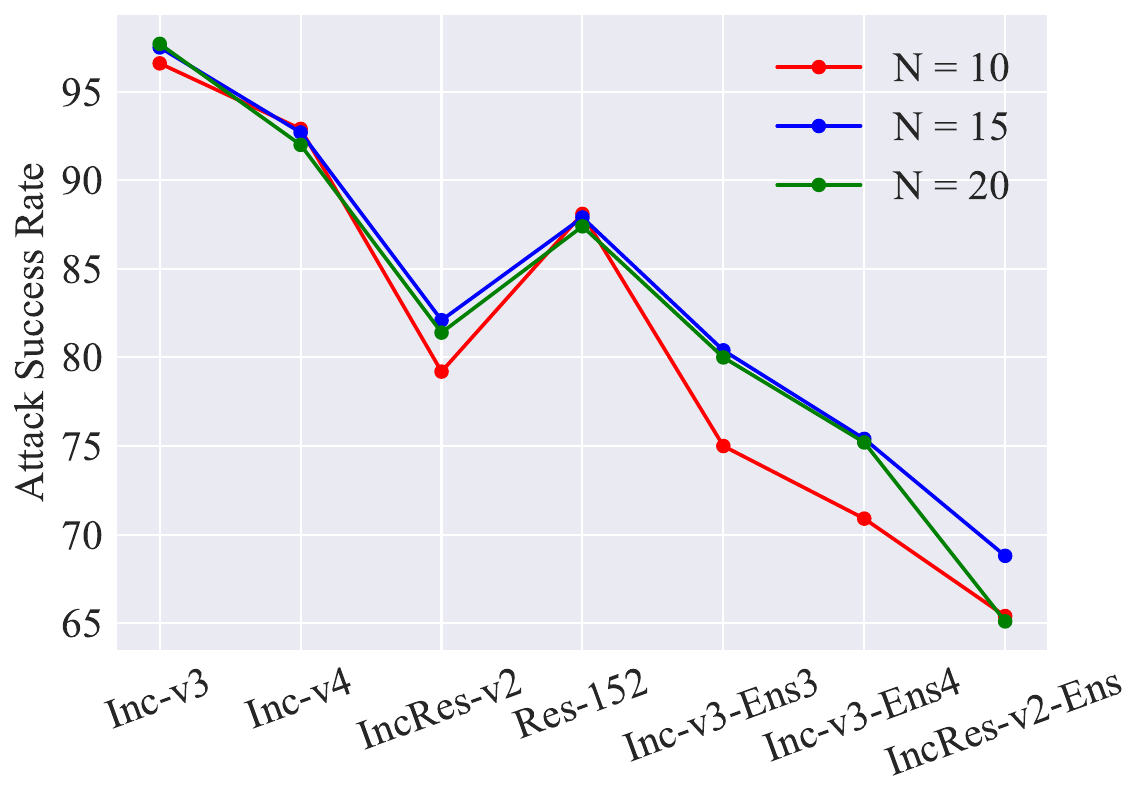}
  \caption{different N}
  \label{fig:N}
\end{subfigure}
\begin{subfigure}{0.23\textwidth}
  \includegraphics[width=\linewidth]{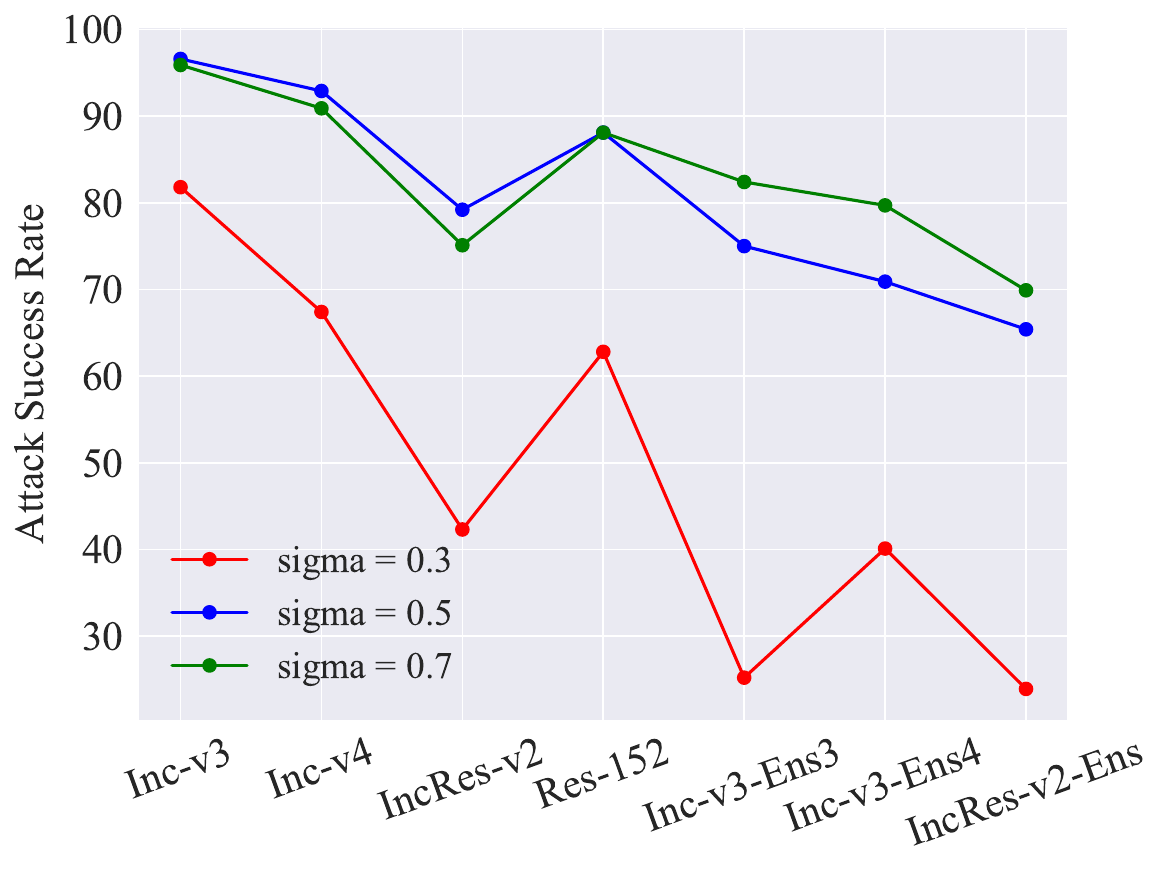}
  \caption{different \textit{sigma}}
  \label{fig:rho}
\end{subfigure}
\begin{subfigure}{0.23\textwidth}
  \includegraphics[width=\linewidth]{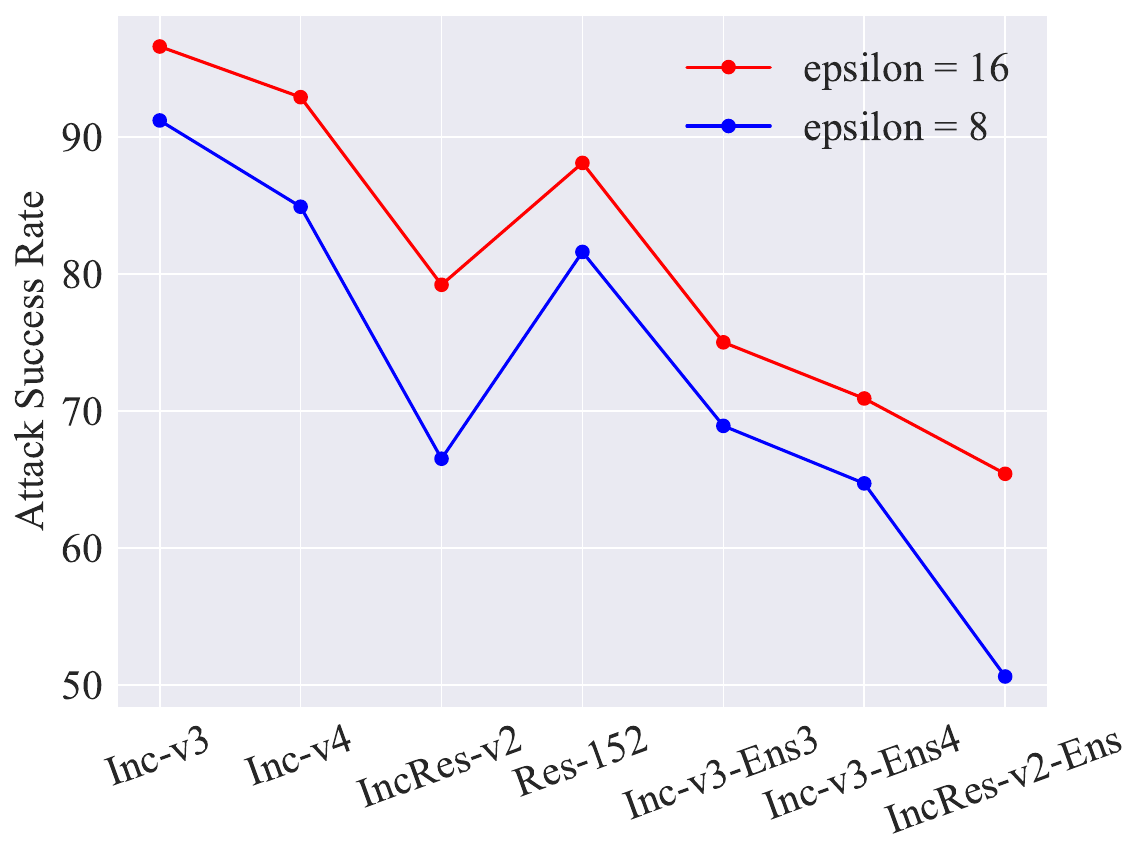}
  \caption{different \textit{epsilon}}
  \label{fig:Sigma}
\end{subfigure}

\caption{GE-AdvGAN attack success rate with different parameters}
\end{figure*}

In this section, we utilise Inception-v3 as the source model to investigate the impact of different \textit{adv lambda}, $N$, \textit{sigma} ($\sigma$), and \textit{epsilon} ($\epsilon$) on the performance of GE-AdvGAN. The remaining parameters are the same as in the main experiments.

\subsubsection{The effect of parameter \textit{adv lambda}}


We use \textit{adv lambda} to control the importance of adversarial loss. A larger value of \textit{adv lambda} indicates higher importance. In Fig.~\ref{fig:adv}, we initially set $N$ to 10, \textit{sigma} to 0.5, and \textit{epsilon} to 16. Subsequently, we tune the \textit{adv lambda} parameter to values of 5, 10, and 15. It can be observed that when \textit{adv lambda} is set to 10, the method exhibits improved attack performance across all models.

\subsubsection{The effect of parameter $N$}


As illustrated in Fig.~\ref{fig:N}, we initially set \textit{adv lambda} to 10, \textit{sigma} to 0.5, and \textit{epsilon} to 16. We then adjust the $N$ parameter to values of 10, 15, and 20. When $N$ is set to 10, the performance noticeably drops compared to the other two values. However, when $N$ is 15 and 20, the performance becomes comparable and competitive across all models except for the last model of IncRes-v2-Ens. Notably, when $N$ is set to 15, the method exhibits superior performance on the IncRes-v2-Ens model.

\subsubsection{The effect of parameter \textit{sigma}($\sigma$)}


As depicted in Fig.~\ref{fig:rho}, we initially set \textit{adv lambda} and $N$ to 10, and \textit{epsilon} to 16. We then change the \textit{sigma} parameter to values of 0.3, 0.5, and 0.7. It can be observed that when \textit{sigma} is 0.3, the performance significantly is the worst. When \textit{sigma} is set to 0.5, the method exhibits better performance on models that had not implemented adversarial training. However, when \textit{sigma} is set to 0.7, GE-AdvGAN demonstrates superior performance on models that implement defensive training.

\subsubsection{The effect of parameter \textit{epsilon} ($\epsilon$)}


As shown in Fig.~\ref{fig:Sigma}, we have set \textit{adv lambda} and $N$ to 10, and \textit{sigma} to 0.5. We then evaluate the \textit{epsilon} parameter with two values, namely 8 and 16. It can be observed that, when \textit{epsilon} is set to 16, GE-AdvGAN achieves the best performance across all models.

\section{Conclusion}
In this work, we propose a novel method called GE-AdvGAN to optimise the training process of the generator parameters. By incorporating frequency domain exploration to determine the direction of gradient editing operation, GE-AdvGAN enables the generation of highly transferable adversarial samples while significantly reducing inference time in comparison with various existing state-of-the-art transferable adversarial attack methods. Extensive experiments conducted on large-scale datasets have evidently demonstrated the superiority of our algorithm. We have open-sourced our replication package of GE-AdvGAN. We anticipate this work provides some insights towards improving the adversarial sample transferability and efficiency against black-box adversarial attack scenarios, paving the way for future research and improvement in the community. 


\bibliography{example}
\bibliographystyle{IEEEtran}

\appendix



\section{Review and analysis of AdvGAN}
As an important component of adversarial generative models, AdvGAN~\cite{xiao2018generating} first uses the generator $G$ to generate perturbations $G(x)$ for the original input $x$. The synthesized adversarial sample $x + G(x)$ is then fed into the Discriminator $D$, which is used to distinguish between adversarial samples and real samples. The goal of $G$ is to generate samples that can deceive the discriminator into classifying them as real samples, while $D$ aims to correctly differentiate between real and adversarial samples. Through iterative and competitive training of $G$ and $D$, AdvGAN ultimately obtains convincing adversarial samples to ensure the success of the attack on the target neural network $f$. It is worth noting that once $G$ is trained, it does not require additional access to the original target network $f$ to generate perturbations for any input data and conduct semi-whitebox attacks. 
\subsection{Loss function of Generator}\label{lfg}
Assuming a target attack scenario (where the label of the benign sample is $b$), in order to maximize the misclassification of the manipulated sample's label by the target model $f$ as the target label $t$, AdvGAN utilizes the loss function $L_{adv}^{f}$ to estimate the likelihood of misleading $f$. The mathematical expression of $L_{adv}^{f}$ is as follows:
\begin{equation}
\label{ladvf}
\small
    L_{adv}^{f}=E_{x}l_{f}(x+G(x),t)
\end{equation}
where $E_{x}$ denotes the expectation value of the input data $x$, according to the unknown distribution $P_{data}$. $l_{f}$ represents the loss function (e.g., cross-entropy loss) used for training the target model $f$. By continuously optimizing and minimizing $L_{adv}^{f}$, we can obtain the adversarial sample whose label is closest to the benign label $b$. At this point, it can be considered that $G$ has been completely trained. Moreover, AdvGAN can also perform untargeted attacks by maximizing the distance between the predicted label and the benign label.
\subsection{Loss function of Discriminator}\label{lfd}
For the discriminator part, AdvGAN utilizes the loss function $L_{GAN}$ to measure the similarity between the manipulated data and the real data in $D$. The mathematical expression of $L_{GAN}$ is as follows:
\begin{equation}
\small
\label{lgan}
    L_{GAN}=E_{x}logD(x)+E_{x}log(1-D(x+G(x)))
\end{equation}
where $E_{x}logD(x)$ measures the discriminator's ability to accurately predict original samples, expecting the prediction to be close to 1. Similarly, $Exlog(1 - D(G(x)))$ assesses the discriminator's inability to accurately predict generated samples $x+G(x)$, hoping that the prediction is close to 0. Therefore, by maximizing the value of $L_{GAN}$, the trained $D$ can make the original samples indistinguishable from the adversarial samples.

In order to maximize the discrimination between generated and real samples by $D$ and to bound the magnitude of the perturbation, AdvGAN incorporates the soft hinge loss, building upon some previous research~\cite{carlini2017towards,liu2016delving,isola2017image}.
\begin{equation}
\small
\label{lhinge}
    L_{hinge}=E_{x}max(0,\left \| G(x) \right \|_{2}-c )
\end{equation}
where $\left \| \cdot \right \|_{2}$ denotes the $L_2$ norm. $c$ represents a user-specified bound.
\subsection{Objective loss function of AdvGAN}\label{eqobjective}
Considering Eq.~\ref{ladvf}-\ref{lhinge} comprehensively, the objective loss function of AdvGAN can be expressed as $L$.
\begin{equation}
\small
    L=L_{adv}^{f}+\alpha L_{GAN}+\beta L_{hinge}
\end{equation}
where $\alpha$ and $\beta$ are hyperparameters that control the relative importance of $L_{GAN}$ and $L_{hinge}$. As we analyzed in Section.~\ref{lfg} and \ref{lfd}, we can generate adversarial samples with different labels and similar appearance to the original samples by taking the extreme value of $L_{adv}^{f}$ and $L_{GAN}$. Therefore, the final trained $G$ and $D$ can be obtained by the following optimization formula:
\begin{equation}
\small
    arg\min_{G}\max_{D}L 
\end{equation}



\end{document}